# A Neural Approach to Language Variety Translation

**Marta R. Costa-jussà[1], Marcos Zampieri[2], Santanu Pal[3]**
[1]TALP Research Center, Universitat Politècnica de Catalunya, Barcelona, Spain
[2]Research Group in Computational Linguistics, University of Wolverhampton, U.K.
[3]Department of Language Science and Technology, Saarland University, Germany
`marta.ruiz@upc.edu`

## Abstract

In this paper we present the first neural-based machine translation system trained to translate between standard national varieties of the same language. We take the pair Brazilian - European Portuguese as an example and compare the performance of this method to a phrase-based statistical machine translation system. We report a performance improvement of 0.9 BLEU points in translating from European to Brazilian Portuguese and 0.2 BLEU points when translating in the opposite direction. We also carried out a human evaluation experiment with native speakers of Brazilian Portuguese which indicates that humans prefer the output produced by the neural-based system in comparison to the statistical system.

## 1 Introduction

In the last five years Neural Machine Translation (NMT) has evolved from a new and promising paradigm in Machine Translation (MT) to an established state-of-the-art technology. A few studies pose that performance difference between Statistical Machine Translation (SMT) and NMT is not as a great as one could imagine (Castilho et al., 2017) while others show interesting challenges for NMT (compared to SMT) such as learning with limited amount of data, out-of-domain, long sentences, low frequency words or lack of word alignment model (Koehn and Knowles, 2017). Even so, NMT systems have constantly ranked in the top positions in the competitions held in MT conferences and workshops such as WMT (Bojar et al., 2016) and WAT (Nakazawa et al., 2016). They have also been achieving commercial success (e.g. Google's GNMT (Wu et al., 2016)).

Far from being settled, the architecture of NMT systems is constantly evolving. Given the youth of the paradigm and while the main structure of encoder-decoder is still maintained, the implementation of such is done either using recurrent neural networks (RNN) with attention mechanisms (Bahdanau et al., 2015), to convolutional neural networks (CNN) (Gehring et al., 2017) and to only attention mechanisms (Vaswani et al., 2017). For the same reason, research in NMT goes in many directions, including minimal units (Sennrich et al., 2016), unsupervised training and low resources (Artetxe et al., 2018) or transfer learning (Zoph et al., 2016), to name and cite just a few.

In this paper we tackle an under-explored problem and apply NMT techniques to translate between language varieties. In previous work (Costa-jussà, 2017), NMT has been used to translate between Spanish and Catalan, two closely-related Romance languages from the Iberian peninsula, outperforming phrase-based SMT approaches. In this paper we test whether this is also true for national varieties of the same language taking Brazilian and European Portuguese as a case study. To the best of our knowledge the use of NMT to translate between national language varieties has not yet been studied and this paper contributes to opening new avenues for future research.

## 1.1 Linguistic Motivation

Translating[1] between varieties of pluricentric languages such as French, Portuguese, and Spanish is an important task carried out in localization companies and language service providers. In the case of Portuguese, although mutually intelligible, the varieties spoken in Brazil and Portugal differ substantially in terms of phonetics, syntax, and lexicon. In previous work, researchers have shown that texts from these two varieties can be discriminated using word and character n-gram-based models with 99% accuracy (Zampieri and Gebre, 2012).

Books written in Brazil are often edited when published in Portugal and vice-versa. Many companies adapt content, product instructions, and user manuals from one country to the other to meet the expectations of costumers and readers of the target country. Brazilian and European Portuguese – henceforth BP and EP – is therefore a particularly relevant and interesting language variety pair to investigate the use of NMT approaches.

A few of the most distinctive characteristics of these two language varieties are summarized next:

- **Orthographic differences:**[2] In the European variety muted consonants such as in *acto, baptismo* (EP) for *act, baptism* (EN) continue to be written whereas in the Brazilian variety for many decades speakers no longer write them *ato, batismo* (BP).

- **Verb tenses:** When expressing progressive events (e.g. *I am running* (EN)), the Brazilian variety prefers the gerund form *correndo* (BP) whereas the European one typically favors the use of the infinitive form *a correr* (EP).

- **Clitics:** Personal pronouns are used in different positions in the two varieties. European Portuguese speakers prefer the use of enclitic personal pronouns, that is after the verb, such as in *Ele viu-me* (EP) whereas Brazilian speakers would typically prefer to use personal pronouns in a proclitic position, before the verb, as in *Ele me viu* (BP) (*He saw me* (EN)).

- **Use of pronouns:** The use of the second person singular pronoun *tu* (*you* (EN)), is only restricted to regional use in Brazil but widespread in Portugal. BP speakers use *você* in most contexts whereas this form is considered to be a formal register in Portugal used only in specific formal contexts similar to the use of *tu - usted* in Spanish and *du - Sie* in German.

Apart from the aforementioned differences, lexical variation is abundant between these two varieties. Examples of these are preferences such as *nomeadamente* which is rare in BP and very frequent in EP and false cognates such as the word *propina*. Its most frequent sense in BP is *bribe* and in EP is *fee*.

## 2 Related Work

There have been a few papers published on translating texts between Brazilian and European Portuguese. One example is the work by Marujo et al. (2011) which proposed a rule-based system to adapt texts from BP to EP. Marujo et al. (2011) used comparable journalistic corpora available at Linguateca (Santos, 2014), namely CETEMPublico and CETEMFolha, a collection of texts collected from the Brazilian newspaper *Zero Hora*, and Ted Talks to evaluate their method.

Another example of a system developed to translate between Brazilian and European Portuguese is the one by Fancellu et al. (2014) who presented and SMT system trained on a parallel collection from Intel translation memories. The authors report 0.589 BLEU score using a Moses baseline system.

Apart from the two aforementioned studies on translating between Portuguese varieties there have been a few studies published on translating between similar languages, language varieties, and dialects of other languages. Examples of such studies include Zhang (1998) on Mandarin and Cantonese Chinese, Scannell (2006) on Irish and Scottish Gaelic, (Goyal and Lehal, 2010) on Hindi and Punjabi, a few studies on Afrikaans and Dutch (Van Huyssteen and Pilon, 2009; Otte and Tyers, 2011), and Hassani (2017) on Kurdish dialects.

---

[1]In this paper, when talking about language varieties, we use the verbs *adapt*, *edit*, and *translate* interchangeably. In previous work Marujo et al. (2011) used the word *adaptation* whereas Fancellu et al. (2014) used the word *conversion*. We consider it, however, as a full-fledged translation task and approach the task as such.

[2]The 1990 orthographic agreement has been recently introduced in both countries diminishing these differences.

To the best of our knowledge, however, the use of NMT is under-explored in these tasks and no language variety translation system has been developed using NMT. The most similar study is the one by (Costa-jussà et al., 2017) who developed a neural-based MT system to translate between Catalan and Spanish. The use of NMT to translate between language varieties is the main contribution of our work.

## 3 Methods

To be able to compare MT approaches, we trained SMT and NMT systems using the same dataset described in Section 3.1. The two systems are described in detail in Section 3.2.

Systems within the SMT category use statistical techniques to compose the final translation. There are a variety of alternatives that are state-of the art, including: n-gram (Mariño et al., 2006), syntax (Yamada and Knight, 2001) or hierarchical to name a few. In this paper, we are using the popular phrase-based system (Koehn et al., 2003).

Systems within the NMT category use a machine learning architecture based on neural networks to compose the final translation. As mentioned, there are several architectures which have been proven state-of-the-art, all of them based on an encoder-decoder schema but using either recurrent neural networks (Cho et al., 2014), convolutional neural networks (Gehring et al., 2017) or the transformer architecture based only on attention-based mechanisms (Vaswani et al., 2017). These architectures can be adapted to deal with different input representations either words, subwords (Sennrich et al., 2016), characters (Costa-jussà and Fonollosa, 2016; Lee et al., 2017) or bytes (Costa-jussà et al., 2017).

In this paper, we are using the first option of recurrent neural networks with an added attention-based mechanism (Bahdanau et al., 2015) and bytes as input representations (Costa-jussà et al., 2017).

### 3.1 Data

Compiling suitable parallel language variety corpora for NLP tasks is not trivial. Popular and freely available data sources (e.g. Wikipedia) used in NLP do not account for regional variation. One possible data source that includes national varieties of the same language are technical user manuals which are often localized between countries. However, user manuals contain a very specific technical language with short and idiomatic sentences representing commands.

We searched for suitable datasets and we acquired an aligned Brazilian - European Portuguese parallel corpus of film subtitle dialogues from Open Subtitles available at Opus[3] (Tiedemann, 2012). We removed all XML tags available in the data. The cleaned corpus, which we will be making available for the community as another contribution of our work[4], comprises 4.3 million sentences in each language for training, with over 33 million tokens for BP and over 34 million tokens for EP. Finally, 2,000 parallel sentences were kept for development and another 2,000 sentences for testing.

### 3.2 Systems

**Statistical-based.** In a phrase-based system, the main model, which is the translation model, is extracted by statistical co-occurrences from a parallel corpus at the level of sentences. This translation model is combined in the decoder with other models to compose the most probable translation given a source input. We built a standard phrase-based system with Moses open source toolkit (Koehn et al., 2007). The main parameters of our implementation include: grow-diagonal-final-and word alignment symmetrization, lexicalized reordering, relative frequencies (conditional and posterior probabilities) with phrase discounting, lexical weights, phrase bonus, accepting phrases up to length 10, 5-gram language model with Kneser-Ney smoothing, word bonus and MERT (Minimum Error Rate Training) optimisation. These parameters are taken from previous work (Costa-jussà et al., 2017).

---
[3]http://opus.lingfil.uu.se/
[4]The clean version of the corpus is available upon request.

**Neural-based.** Specifically, neural MT computes the conditional probability of the target sentence given the source sentence by means of an encoder-decoder or sequence-to-sequence (seq2seq) architecture, where the encoder reads the source sentence, does a word embedding, and encodes it into an intermediate representation using a bidirectional recurrent neural network with Long Short Term Memory units (LSTM) as activation functions. Then, the decoder, which is also a recurrent neural network, generates translation based on this intermediate representation. This baseline seq2seq architecture is improved with an attention-based mechanism (Bahdanau et al., 2015) which allows for the introduction of contextual information while decoding. This architecture is extended to deal with bytes (Costa-jussà et al., 2017) to overcome unknown words. We are adopting the same parameters as set in mentioned previous work. We use an in-house Theano implementation based on code available[5]. We use a bidirectional LSTM of 512 units for encoding, a batch size of 32, no dropout and ADAM optimization. For more details about the architecture refer to previous work (Costa-jussà et al., 2017).

### 3.3 Human Evaluation

To validate the results obtained with the automatic evaluation metrics, we used the ranking and rating features available in CATaLog online (Pal et al., 2016a; Pal et al., 2016b), a web-based CAT tool developed for translation process research. We ask native speakers of Brazilian Portuguese first to compare segments translated by NMT and SMT, choosing the best output, and subsequently to rate translations taking both fluency and adequacy into account using a 1 to 7 Likert scale. More information and the results of these experiments are presented in Section 4.2.

## 4 Results

### 4.1 Automatic Metrics

In this section we present the results obtained by the statistical-based system based of phrases and the neural-based system based on seq2seq with attention and bytes in terms of BLEU score (Papineni et al., 2002). Table 1 presents the results obtained by the three systems when translating from EP to BP and Table 2 presents results obtained from BP to EP. The best results for each setting are presented in bold.

| System | BLEU Score |
| --- | --- |
| Phrase-based SMT | 47.68 |
| Neural MT | **48.58** |

Table 1: European to Brazilian Portuguese translation results in terms of BLEU score.

| System | BLEU Score |
| --- | --- |
| Phrase-based SMT | 47.34 |
| Neural MT | **47.54** |

Table 2: Brazilian to European Portuguese translation results in terms of BLEU score.

We observed that in both directions the NMT system outperformed the SMT approach. The neural system obtained the best performance translating from European to Brazilian Portuguese achieving 48.58 BLEU points on average, and from Brazilian to European Portuguese achieving 47.54 BLEU points on average. This performance was 0.9 and 0.2 BLEU points better than the average performance obtained by the SMT system.

---

[5] https://github.com/nyu-dl/dl4mt-c2c

## 4.2 Human Evaluation

Following the best practice in translation evaluation, as observed in the WMT shared tasks (Bojar et al., 2016; Bojar et al., 2017), we present two human evaluation experiments to validate the results obtained with the automatic evaluation metrics. The first one is a ranking experiment in which participants were asked to choose which translation of a given segment they preferred without knowing which system produced the translations. The second one is a pilot rating experiment in which participants were asked to rate translations using a 1 to 7 Likert scale.

We start by reporting the results obtained in the ranking experiment. We evaluated the quality of the EP-BP translation direction asking a group of seven native speakers of BP to rank sentences produced by the SMT system and by the NMT system. We presented native speakers with the source segment in EP and two translations in BP. Their task was to select the best output (ties were allowed). The decision to use only BP native speakers was motivated by the findings reported in (Goutte et al., 2016) which indicate that speakers of BP are generally not familiar with what is acceptable in the European variety and vice-versa.

We considered the full set of 2,000 test sentences and disregard sentences where 1) no transformation has been made (the source, outputs, and the reference are the same), and 2) transformations have been made but NMT and SMT outputs are the same. After filtering, 679 distinct segments were left for the human evaluation. We randomly selected 20% of these segments 679 (136 sentences) and presented them to two annotators to calculate inter-annotator agreement. Finally, the set of 815 segments, 679 segments plus 136 segments (20% redundancy), divided into 23 sub-sets and presented to each of the annotators. Each annotator evaluated between three and four sub-sets.

To assess the reliability of the rankings, we first compute the agreement between the annotators. We report substantial inter-annotator agreement achieving 0.88 pairwise Kappa score. We present the results obtained by the ranking experiment in terms of the percentage of segments in which 1) NMT was preferred by the annotator, 2) the two segments were consider the same, 3) SMT was preferred by the annotator. Results are summarized in Table 3. We observed that the NMT output was judged to be equal or better the SMT output in 57.81% of the cases. The NMT output was preferred in 48.43% of the rankings and judged to be of same quality as SMT in 9.38% of the cases.

| Outcome | Percentage of Cases |
|---|---|
| NMT preferred | 48.43% |
| Ties | 9.38% |
| SMT preferred | 42.19% |

Table 3: Human evaluation scores - Raking experiment.

To further investigate how humans perceive the quality of the EP-BP translations, we carry out a pilot rating experiment with two of the aforementioned annotators. From the 679 segments included in the ranking experiment, we randomly selected 100 segments and create two sub-sets of 50 segments each. We provide each annotator with a sub-set and ask them to rate the translations taking both fluency and adequacy into account using a Likert scale of 1 to 7. The average results obtained by each system in the pilot rating experiment are presented in Table 4.

| Outcome | Average Score |
|---|---|
| Neural MT | 5.4 |
| Phrase-based SMT | 5.1 |

Table 4: Human evaluation scores - Ranking pilot experiment.

The average results suggest that humans have a rather positive opinion about the translations produced by both systems. The average score obtained by the NMT system was 5.4 whereas the SMT obtained 5.1. The outcomes of both the ranking and the rating experiments confirm that the NMT system produces a higher quality output than SMT for this task and that the improvement in BLEU score is indeed perceived by humans. We are currently replicating this pilot experiment with more annotators.

### 4.3 Discussion

We observed several interesting patterns in the NMT and SMT outputs and here we present a brief linguistic analysis of the output of both systems.

In our analysis, we observed that NMT was generally better in handling the transformation from the proclitic pronominal position generally used in BP to the enclitic pronominal position generally preferred in EP as shown in Example 1. This is just true for declarative-affirmative main clauses (other syntactic contexts involve proclisis).

(1) **SOURCE (EP)**: Fizeste-me chorar
    **SMT**: Fez chorar.
    **NMT:** Você me fez chorar.
    **REFERENCE (BP):** Me fez chorar.
    **EN:** You made me cry.

In Example 1 NMT produces the correct translation in a particularly challenging case in which *tu* was omitted in the source segment. Even when this re-ordering operation had to occur in the opposite direction, EP enclitic and BP proclitic, NMT was able to handle it better than SMT as demonstrated in Example 2.

(2) **SOURCE (EP)**: Nem me estás a ouvir, querido.
    **SMT**: Nem me está a ouvir, querido.
    **NMT:** Nem está me ouvindo, querido.
    **REFERENCE (BP):** Nem está me ouvindo, querido.
    **EN:** You are not even hearing me, honey.

NMT also proved to be better in translating the infinitive form used in EP *a ouvir* to the gerund preferred in BP *ouvindo* (EN: *doing*). Example 2 is a good example in which NMT produced a better output than SMT handling both pronominal positions and verb form correctly.

In Example 1 both systems translated *fizeste* from the second person singular *tu* preferred in EP to the third person singular *fez* as in *você fez* (EN: *you did*) preferred in BP. However, we observed a wide variation when handling the forms *tu* and *você* and their respective verbs. We spotted several translations in which SMT handled this phenomenon better than the NMT system as in Example 3.

(3) **SOURCE (EP)**: Tu não pedes nada a ninguém.
    **SMT**: Você não pede nada a ninguém.
    **NMT:** Tu não pedes nada a ninguém.
    **REFERENCE (BP):** E você jamais pediria nada, não é?
    **EN:** You don't ask anything to anyone.

Finally, it is important to note that, as evidenced in Example 3, in many cases the reference BP translation was substantially or completely different from the EP source. This is probably related to the data domain, movie subtitles, which presents a great deal of variability and encourages creative translations, in this case from the language of the movie to the respective variety of Portuguese. This variability certainly led both systems to obtain lower BLEU scores. Nevertheless, the results of the pilot rating experiment indicate that humans have a rather positive opinion about both systems' output and that NMT is preferred.

## 5 Conclusion and Future Work

This paper presented the first NMT system trained to translate between national language varieties. We used the language variety pair Brazilian and European Portuguese as an example and a parallel corpus of subtitles to train the NMT system.

Compared to an SMT system trained on the same data, we report a performance improvement of 0.9 BLEU points in translating from European to Brazilian Portuguese and 0.2 BLEU points when translating in the opposite direction. Our results indicate that the NMT system produces better translations than the SMT system not only in terms of BLEU scores but also according to the judgments of seven native speakers. The human evaluation experiments contribute to recent empirical evaluations on the quality of NMT output beyond automatic metrics (Castilho et al., 2017).

We would like to carry out a more comprehensive human evaluation experiment in future work. This includes a rating experiment with more annotators and replicating the EP to BP evaluation to the BP to EP translations. A pilot experiment with EP speakers indicates that NMT output is also preferred. We would like to measure the cognitive effort and post-editing time for both SMT and NMT outputs. This could provide valuable information for translation and localization companies about translation productivity. Finally, and since NMT is progressing fast, future comparisons will be built with other NMT architectures like (Vaswani et al., 2017).

## Acknowledgments

We would like to thank the annotators who helped us with the human evaluation experiments. This work is supported in part by the Spanish Ministerio de Economía y Competitividad, the European Regional Development Fund and the Agencia Estatal de Investigación, through the postdoctoral senior grant Ramón y Cajal, the contract TEC2015-69266-P (MINECO/FEDER,EU) and the contract PCIN-2017-079 (AEI/MINECO).